\documentclass[conference,a4paper]{APSIPA2017}
\usepackage{multirow}
\usepackage[psamsfonts]{amssymb}
\usepackage{amsxtra}
\usepackage{threeparttable}
\usepackage{multicol}
\usepackage{times}
\usepackage{epsfig}
\usepackage{amsmath}
\usepackage{amssymb}
\usepackage{algorithmic}
\usepackage[lined,boxed,commentsnumbered]{algorithm2e}
\usepackage{bm}
\usepackage{caption}
\usepackage{cite}
\usepackage{color}
\usepackage[export]{adjustbox}
\usepackage{graphicx}
\usepackage{subcaption}
\usepackage{url}
\usepackage[bookmarks=false]{hyperref}
\graphicspath{{./figs/} {./figs/scene_01/} {./figs/scene_02/} 
}

\begin{document}

\title{Tracking objects using 3D object proposals}

\author{%
\authorblockN{%
Ramanpreet Singh Pahwa\authorrefmark{1},
Tian Tsong Ng\authorrefmark{2} and
Minh N. Do\authorrefmark{3}
}
\authorblockA{%
\authorrefmark{1}
Institute for Infocomm Research, Singapore \\
E-mail: ramanpreet$\_$pahwa@i2r.a-star.edu.sg  Tel: +65-64082441}
\authorblockA{%
\authorrefmark{2}
Institute for Infocomm Research, Singapore \\
E-mail: ttng@i2r.a-star.edu.sg  Tel: +65-64082517}
\authorblockA{%
\authorrefmark{3}
University of Illinois at Urbana-Champaign \\
E-mail: minhdo@uiuc.edu  Tel: +1-217-244-4782}%
}

\maketitle
\thispagestyle{empty}

\begin{abstract}
3D object proposals, quickly detected regions in a 3D scene that likely contain an object of interest, are an effective approach to improve the computational efficiency and accuracy of the object detection framework. In this work, we propose a novel online method that uses our previously developed 3D object proposals, in a RGB-D video sequence, to match and track static objects in the scene using shape matching. Our main observation is that depth images provide important information about the geometry of the scene that is often ignored in object matching techniques. Our method takes less than a second in MATLAB on the UW-RGBD scene dataset on a single thread CPU and thus, has potential to be used in low-power chips in Unmanned Aerial Vehicles (UAVs), quadcopters, and drones.
\end{abstract}

\begin{section}{Introduction}
The rapid development of low-powered Unmanned Aerial Vehicles (UAVs), drones and service robots has introduced a need for automatic detection of interesting objects present in a scene. Such applications may not only assist in navigation of these devices but also help in localizing, tracking and identifying the objects that are present in a scene. This paper presents a framework where we leverage on color and depth information, per frame, to obtain a global heatmap of the scene. This heatmap is used to find interesting $3$D objects in the scene. Using RGB-D SLAM enables us to locate these $3$D objects in the scene. We use Jaccard index to match objects in the scene. We show that using a simple Intersection over Union (IoU) is much faster than using complex techniques such as $3$D feature matching for static objects in the scene, while maintaining similar accuracy.

The popularity of depth cameras in scientific community has led to explosion of recent discoveries of various applications in computer and robot vision \cite{wilson2010using,  shotton2013real, Biswas, Song_TCSVt2014,raman_icip2014}, including object recognition \cite{depth_obj_recog}. While $3$D object proposals are a relatively new idea \cite{raman_tcsvt_2016}, they are heavily inspired from $2$D object proposals \cite{cheng2014bing, CPMC, MCG, zitnick2014edge}. $2$D object proposal techniques have become immensely popular in object detection systems. Instead of finding precise and exclusive boundaries for objects, modern $2$D object proposal techniques quickly identify regions (potentially highly overlapped) that are very likely to contain an object. $3$D object proposals build on this idea. An effective way to obtain true shape, size, and orientation of $3$D objects is to use depth cameras. $3$D object proposals utilize depth information to estimate scene geometry and use the additional information to extend the $2$D objects proposals to $3$D by estimating $3$D cuboids for each object on interest. $3$D proposals are inherently more useful than $2$D object proposals as they provide us with accurate physical dimensions of the objects of interest that are present in the scene which can be used for various tasks such as object manipulation and scene understanding.

In this work, we utilize online $3$D object proposals for RGB-D video input of a static scene with the main focus on matching objects. We use the results obtained per-frame by existing $3$D proposal techniques as an input to our system. We leverage on segmentation cues provided by depth information and aggregate them over consecutive frames in $3$D by estimating the camera poses using RGB-D SLAM. This enables us to obtain accurate $3$D boundaries of the objects of interest in the scene and use these boundaries as cues to match objects in the scene. 

More specifically, this work focuses on indoor, static scenes containing a major supporting plane. However, it can be easily extended to multiple supporting planes. The objects are allowed to occlude each other and partially seen in some images as long as their complete view is covered over a consecutive number of frames. Note that we do not assume their distance to the camera.
We make the following contributions:
\itemize
\item We use the camera pose estimated by a depth based
SLAM technique to efficiently register the frames to a
global point cloud. We constantly update our global $3$D object proposals as the camera moves around in the scene. 
\item We match the current frame's $3$D object proposals to global $3$D object proposals using shape matching to find matching objects in the scene. Any new object seen is added to the global object dataset and tracked from thereon.

Our paper is structured as follows. In Section \ref{sec::related_work}, we review other works that are related to our research topic. In Section \ref{sec::3d_proposals}, we provide a brief overview of $3$D object proposals. Section \ref{sec::shape_matching} presents the proposed $3$D shape matching steps in detail, highlighting our contributions and observations at each stage. We report our results in Section \ref{sec::results}. Finally, in Section \ref{sec::conclusion} we conclude this paper and discuss future research directions.
\end{section}

\begin{section}{Related Work} \label{sec::related_work}
\textbf{Object proposals}: A lot of work has been done recently on $2$D object proposals. Traditionally, sliding windows are used along with a detector to identify objects in a given scene. Many of the state-of-the-art techniques in object detection have started using a generic, object-class agnostic proposal method that finds anywhere between $100$-$10,000$ bounding boxes in an image. These areas are considered to have the maximum likelihood to contain an object in them. Such methods vary widely from using linear classifiers, BING \cite{cheng2014bing}, graph cuts, CPMC \cite{CPMC}, graph cuts with an affinity function \cite{hoiem_PAMI2014}, normalized cuts, MCG \cite{MCG} to using  random forests, and edge-boxes \cite{zitnick2014edge}. Ren \emph{et~al.} \cite{NIPS2015_Faster_RCNN} use deep learning in a supervised manner to find $2$D object proposals and perform object detection simultaneously. 
Unfortunately, these algorithms have limited repeatability \cite{hosang2015makes}. Even changing one pixel exhibits markedly different outcomes. $3$D object proposals \cite{raman_tcsvt_2016} try to overcome these issues by leveraging on depth and temporal information. They use the depth information to filter out majority of redundant object proposals and exploit scene planarity to identify and remove underlying planes in the scene. 

\textbf{Depth based SLAM}: RGB-D SLAM \cite{steinbrucker2011real, scherer2013efficient, engelhard2011real} allows to quickly acquire colored $3$D pointcloud of the indoor scenes with a hand-held RGB-D camera such as Kinect. It uses visual features to identify matching points in acquired images, and uses RANSAC to robustly estimate the $3$D transformation (camera pose) between them.
 
\textbf{$3$D feature matching}: Usually $3$D feature descriptors \cite{sai_3dhop, tombari2013performance, guo_tpami} are used for matching objects in $3$D. However, these methods usually come with an extremely large memory footprint and matching complexity which is not needed for static objects. For example, the three most commonly used $3$D feature descriptors - $SHOT$, $FPFH$, and  $RoPS$ have a dimensionality of $352$, $33$, and $135$ respectively. 
\end{section}

\begin{section}{3D Object Proposals}\label{sec::3d_proposals}
\begin{figure}[t]
\centering
  \includegraphics[width=\columnwidth]{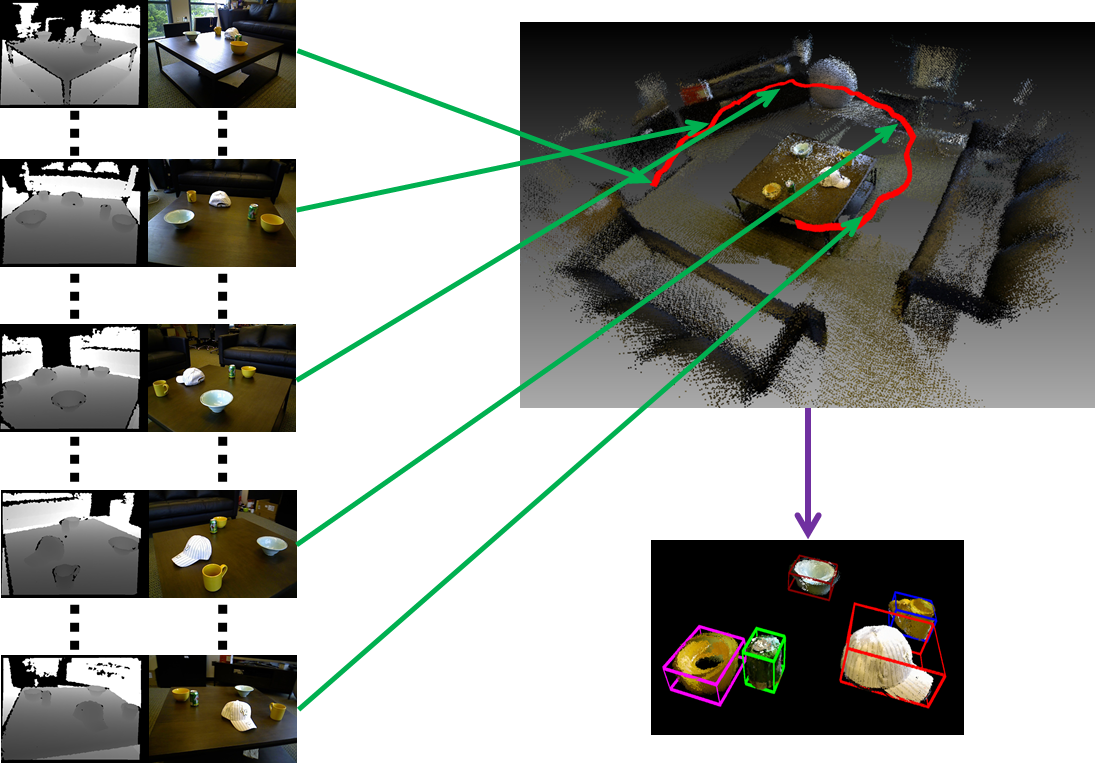}
\caption{Scene data is captured using a RGB-D camera. The aligned color and depth images are used for camera pose estimation (illustrated by the red trajectory). Our approach uses this information along with any generic $2$D object proposals to fuse and filter the data in $3$D and output precise $3$D object proposals (denoted by colored $3$D bounding-boxes).}
\label{fig::our_traj_demo}
\end{figure}
We start by giving an overview of the $3$D object proposals technique \cite{raman_tcsvt_2016}. At high level, the algorithm is designed to fuse the depth information with the generic $2$D object proposals obtained from color images to obtain $3$D object proposals per frame. This enables us to exploit using $3$D geometry of the scene. These object proposals are improved as the camera moves around in the scene. Figure ~\ref{fig::our_traj_demo} illustrates the basic setup of the problem and an example result obtained. 

Before presenting the main components of our approach,
we first introduce the initialization process with $3$D object proposals. We collect $N$ video frames per scene using a RGBD camera. Every $i^{th}$ video frame consists of a color image $\bm{I}_i$, a depth image $\bm{Z}_i$, and the pose of the camera $\bm{P}_i$, using a depth based SLAM method such as Dense Visual SLAM \cite{kerl2013dense}. The camera pose contains the rotation and translation measurements of the camera in the world coordinate frame of reference: $\bm{P}_i$ = [$\bm{R}_i$, $\bm{t}_i$]. The RGB-D camera is assumed to be pre-calibrated. The camera intrinsic parameters - the focal length in x and y directions and optical center are denoted by $f_x$, $f_y$, [$c_x$, $c_y$], respectively. Together, these are represented by the intrinsic calibration matrix $\bm{K}$. We use a generic $3$D object proposal technique \cite{raman_tcsvt_2016} to obtain $M$ $3$D object proposals per frame:
\begin{align} \label{Eq::3d_proposals}
\bm{BB}^{j} &= [x^j,y^j,z^j,l^j,w^j,h^j], \enskip j \in {1, \hdots, M}, \\
\bm{r}^{j} &= [r^j_x,  r^j_y, r^j_z] ^{\intercal}.
\end{align}
where, [$x$, $y$, $z$] denote the top-left $3$D coordinate of the bounding box, and $l$, $w$, and $h$ refer to the length, width and the height in x,y and z directions respectively. $\bm{r}$ refers to the rotation of the bounding box to align it to the horizontal direction of the scene so that the bounding boxes ``rest" on the underlying planes such as table and floor. A few $3$D proposals are shown in Fig.~\ref{fig::our_traj_demo}. These $3$D object proposals per image are treated as an input for our framework. Thus, the dimensionality of our feature descriptor is only of size nine which makes our shape matching extremely fast.
\end{section}

\begin{section}{Shape matching using 3D object proposals}\label{sec::shape_matching}
In this section, we describe how we use $3$D object proposals to identify and label $3$D objects in the scene using shape matching. 

\begin{figure}[t!]
\centering
  \includegraphics[width=\columnwidth]{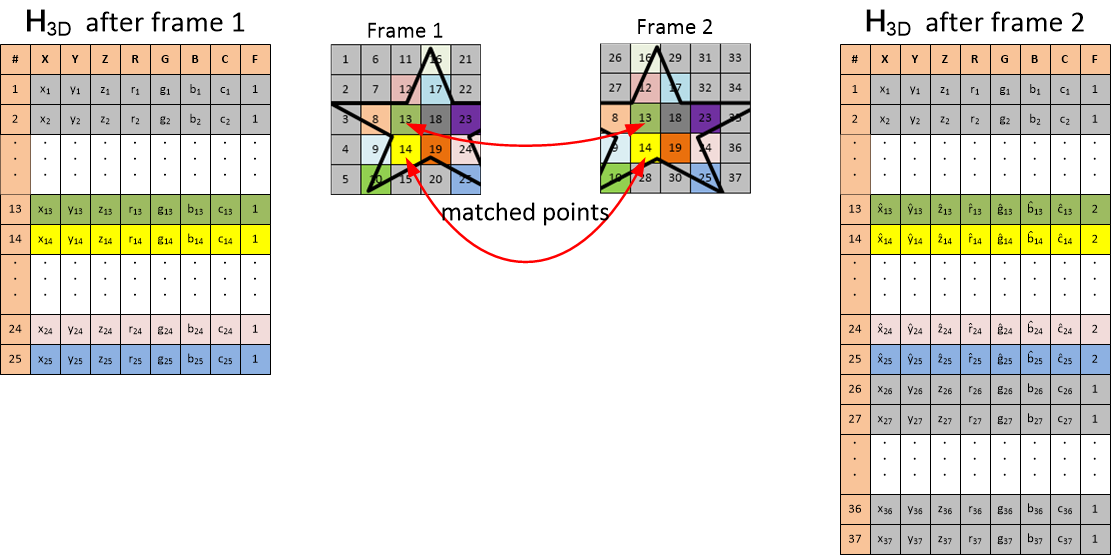}
\caption{We identify matching and non-matching points across consecutive points using RGB-D SLAM. The heat value and average location of the $3$D points representing the matching points is updated after every frame. The non-matching points are identified as new points in the scene and added to the global $3$D heatmap.}
\label{fig::matching_pts}
\end{figure}

Let us assume that we obtain $M$ $3$D objects after processing first frame in a given scene using \cite{raman_tcsvt_2016}. These objects are labelled sequentially from $1$ to $M$. After each frame, we update the $3$D heatmap of the scene as shown in Fig.~\ref{fig::matching_pts}. Every frame update provides us with two new heatmaps - an updated global heatmap and the current frame's heatmap. We use the current heatmap to estimate the $3$D boundaries of objects present in the current frame. As we estimate the camera per frame using RGB-D SLAM, we can transform these object boundaries to the first frame of reference. These object boundaries are compared with the existing database of object boundaries present in the scene. If the new object boundary overlaps significantly with an existing object, it is considered as the same object seen previously and assigned the corresponding label. However, if the object boundary does not overlap with any current existing objects in the scene, it is considered a new object and assigned a new label and added to the existing object database. We use Intersection over Union (IoU) in $3$D to find the overlap value of the $i^{th}$ current $3$D object proposal as follows:
\begin{align}
\text{IoU}(i) = \max_{j}\left(\frac{\text{BB}_\text{c}(i) \cap \text{BB}_\text{o}(j)}{\text{BB}_\text{c}(i)  \cup \text{BB}_\text{o}(j)}\right) \quad \forall j \text{ } \in  \text{ } 1, \hdots , M 
\end{align}
where $\text{BB}_\text{c}(i)$ refers to the $i^{th}$ $3$D bounding box of the current frame's $3$D object proposal and $\text{BB}_\text{o}(j)$ refers to the $j^{th}$ $3$D bounding box of the existing objects in the scene. The intersection and union in $3$D are computed similar to their corresponding $2$D versions as:
{\small
\begin{align}
x^i \cap x^j = \begin{cases} \multirow{2}{*}{0} &
 \text{if  } x^i +l^i < x^j \text{ or} \\ & x^j +l^j < x^i  \\ 
 max(x^j+l^j-x^i, x^i+l^i-x^j) & \text{otherwise}, \end{cases}
\end{align}
\begin{align}
x^i \cup x^j = \begin{cases} \multirow{2}{*}{$l^i + l^j$} &
 \text{if  } x^i +l^i < x^j \text{ or} \\ & x^j +l^j < x^i  \\ 
l^i+l^j - (x^i \cap x^j) & \text{otherwise}, \end{cases}
\end{align}
}

\begin{align}
\text{BB}(i) \cap \text{BB}(j) &= |x^i \cap x^j|.|y^i \cap y^j|.|z^i \cap z^j|\\
\text{BB}(i) \cup \text{BB}(j) &= |x^i \cup x^j|.|y^i \cup y^j|.|z^i \cup z^j|
\end{align}

There are various benefits of using shape matching. It is extremely fast as we only need to compute the $3$D boundaries of objects. The IoU can be computed in $O(M)$, where $M$ is the number of existing objects in the scene. Moreover, we use the actual position of the objects in the scene. If an object is occluded by another object temporarily and seen later, it is immediately re-identified as the original object and assigned the same label as previously. Using $3$D feature matching is slow and sensitive to conditions such as lighting and camera orientation. Meanwhile, shape matching overcomes these issues as it relies on the $3$D pose and location of the objects in the scene.
\end{section}

\begin{section}{Results}\label{sec::results}
We use UW-RGBD scene dataset \cite{lai_icra14} to conduct our experimental analysis and evaluation. The dataset contains $14$ scenes reconstructed from RGB-D video sequences containing furniture and some table-top objects such as caps, cereal boxes and coffee mugs. The scenes contain depth and color frames from a video collected by moving around the scene.  The dataset provides a globally labelled $3$D point cloud. 

After we obtain a $3$D heatmap per frame, we use multi-view information to fuse this information together. We used Dense Visual SLAM \cite{kerl2013dense} to obtain the camera pose per frame to fuse the frames together. The entire process takes $3.03s$ for VGA resolutions on average in MATLAB. However, a majority of the time ($>2.0s$) is spent in storing and accessing the $3$D point cloud. As we aim for a fast and efficient algorithm that is capable of online processing of $3$D object proposals, we downsample the images by $2$. This reduces the time taken per frame to less than one second in MATLAB on a single-core CPU while maintaining a similar accuracy in matching and tracking objects. We obtain on average $6.57$ $3$D object proposals per scene. Our results can be further improved with an improved camera pose estimation, as in some cases the objects break into two or more discontinuous point clusters due to noisy camera poses. This results in multiple distinct object proposals for one object, essentially dividing the object into two or more pieces.

Figures~\ref{fig::uwrgbd_scene01}, \ref{fig::uwrgbd_scene02} show some sample frames in the video sequence. Full videos can be accessed \href{https://goo.gl/WjvCsl}{here}. We observe that our shape matching technique is able to match static objects successfully in most occasions. The white soda can is hiding behind the cereal box and the green soda can is assigned label $5$ in Fig.~\ref{fig::uwrgbd_scene01}(a). Once it is seen in frame $93$ in Fig.~\ref{fig::uwrgbd_scene01}(b), it is assigned label $12$. The green soda can is occluded by cereal box in frame $522$, but as soon as it is observed again, it is reassigned the same label $5$ as shown in Fig.~\ref{fig::uwrgbd_scene01}(f). We observe the same behavior in Fig.~\ref{fig::uwrgbd_scene02} where the green soda can is initially hidden behind cereal box and all the objects occluded by the cereal box are identified as same objects seen before once they reappear in the scene. This shows that, for static objects, we can use an extremely fast shape matching in a RGB-D video sequence rather than expensive feature descriptors \cite{sai_3dhop, tombari2013performance, guo_tpami} that are storage and computationally  intensive.
\begin{figure}[t!]
\centering
	\begin{subfigure}[b]{0.15\textwidth}
	\includegraphics[width=\textwidth]{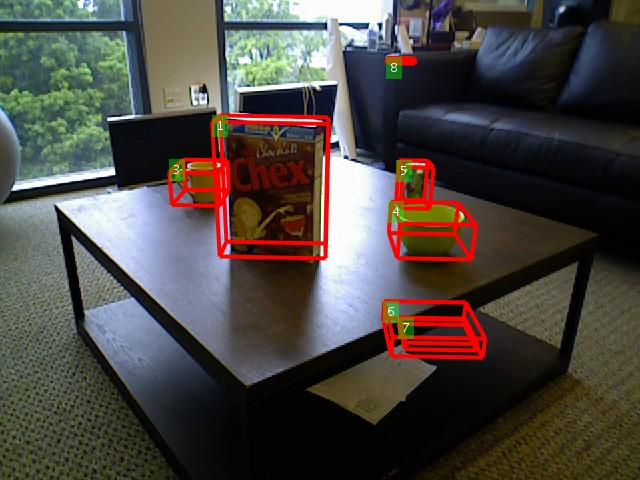}
	\caption{Frame: $2$}
	\end{subfigure}
  \begin{subfigure}[b]{0.15\textwidth}
	\includegraphics[width=\textwidth]{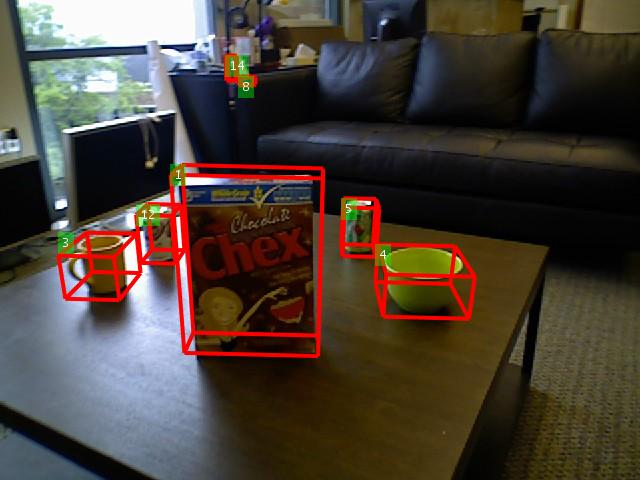}
	\caption{Frame: $93$}
	\end{subfigure}
  \begin{subfigure}[b]{0.15\textwidth}
	\includegraphics[width=\textwidth]{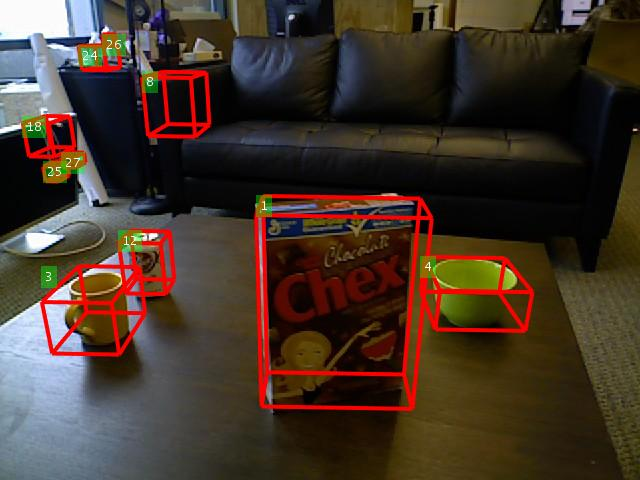}
	\caption{Frame: $148$}
	\end{subfigure}
  \begin{subfigure}[b]{0.15\textwidth}
	\includegraphics[width=\textwidth]{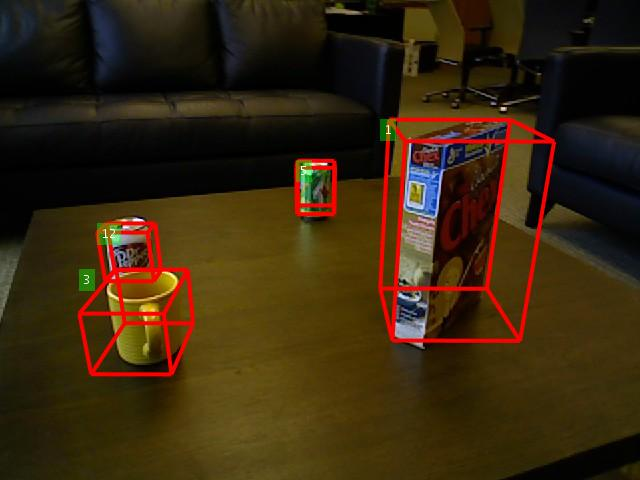}
	\caption{Frame: $248$}
	\end{subfigure}
  \begin{subfigure}[b]{0.15\textwidth}
	\includegraphics[width=\textwidth]{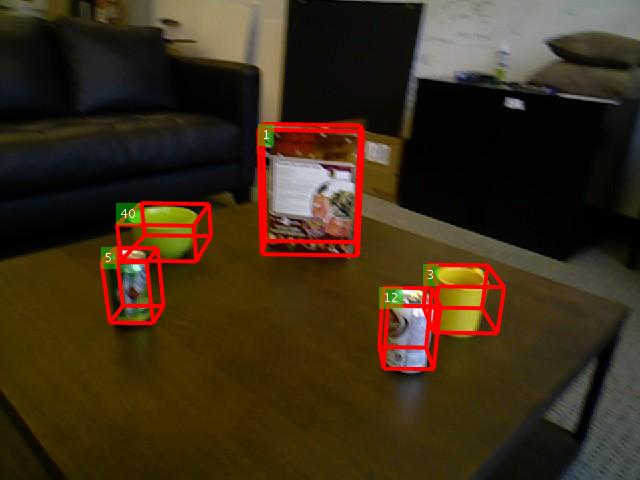}
	\caption{Frame: $522$}
	\end{subfigure}
  \begin{subfigure}[b]{0.15\textwidth}
	\includegraphics[width=\textwidth]{frame_600.png}
	\caption{Frame: $600$}
	\end{subfigure}
  \begin{subfigure}[b]{0.15\textwidth}
	\includegraphics[width=\textwidth]{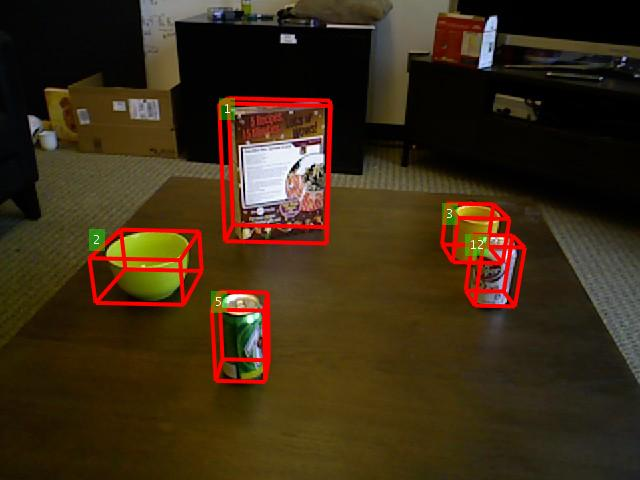}
	\caption{Frame: $616$}
	\end{subfigure}
  \begin{subfigure}[b]{0.15\textwidth}
	\includegraphics[width=\textwidth]{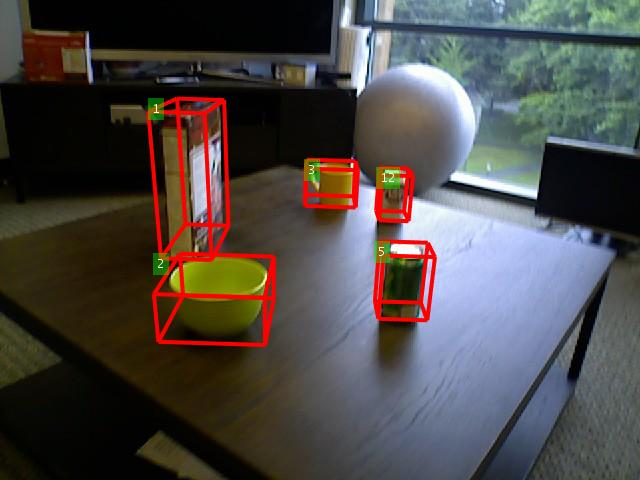}
	\caption{Frame: $768$}
	\end{subfigure}
  \begin{subfigure}[b]{0.15\textwidth}
	\includegraphics[width=\textwidth]{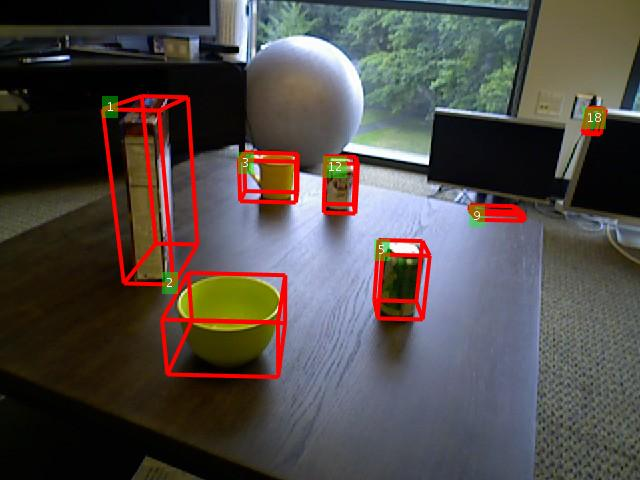}
	\caption{Frame: $834$}
	\end{subfigure}
  \caption{We show our shape matching results for a sample scene in the UW-RGBD scene dataset. The objects maintain the same label throughout the capturing process. The objects occluded by the cereal box retain their label as soon as they are visible again in the scene.}
 \label{fig::uwrgbd_scene01}
\end{figure}

\begin{figure}[t!]
\centering
	\begin{subfigure}[b]{0.15\textwidth}
	\includegraphics[width=\textwidth]{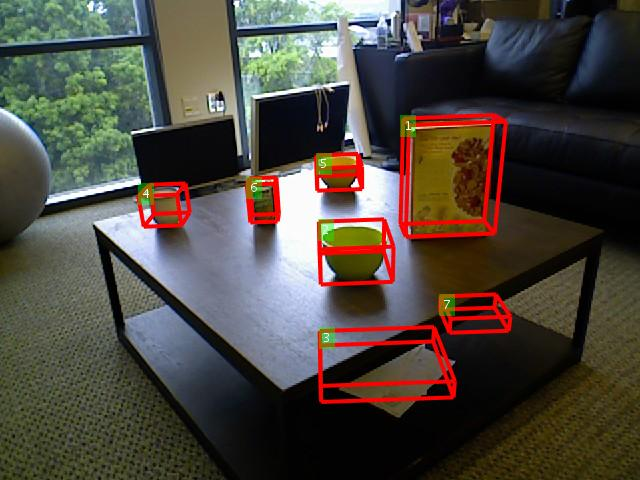}
	\caption{Frame: $1$}
	\end{subfigure}
  \begin{subfigure}[b]{0.15\textwidth}
	\includegraphics[width=\textwidth]{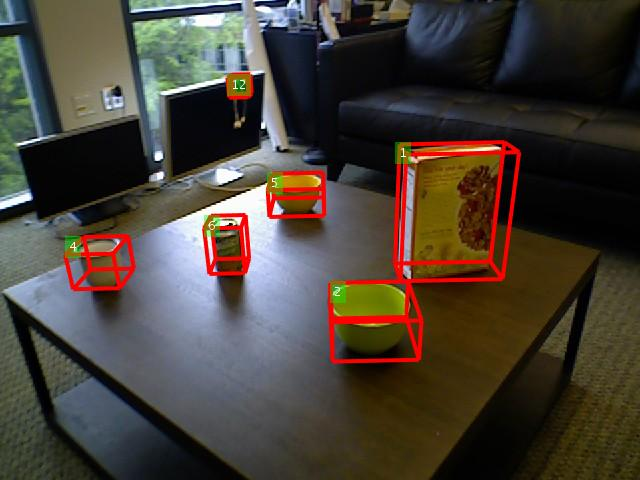}
	\caption{Frame: $100$}
	\end{subfigure}
  \begin{subfigure}[b]{0.15\textwidth}
	\includegraphics[width=\textwidth]{frame_200.png}
	\caption{Frame: $200$}
	\end{subfigure}
  \begin{subfigure}[b]{0.15\textwidth}
	\includegraphics[width=\textwidth]{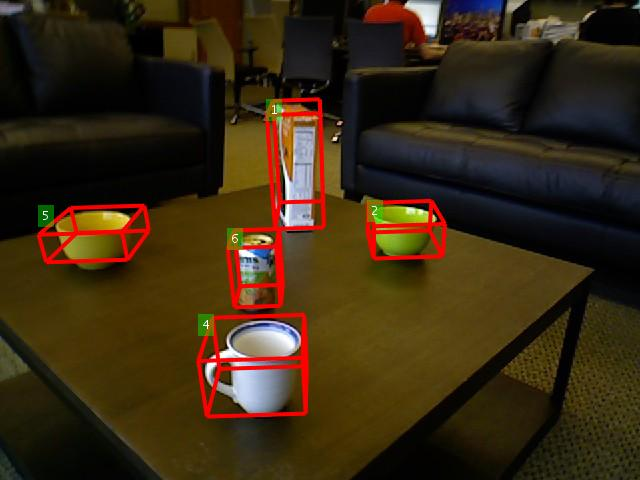}
	\caption{Frame: $300$}
	\end{subfigure}
  \begin{subfigure}[b]{0.15\textwidth}
	\includegraphics[width=\textwidth]{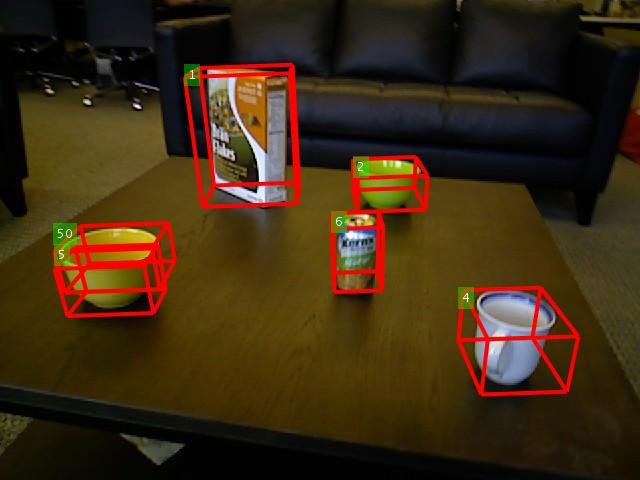}
	\caption{Frame: $394$}
	\end{subfigure}
  \begin{subfigure}[b]{0.15\textwidth}
	\includegraphics[width=\textwidth]{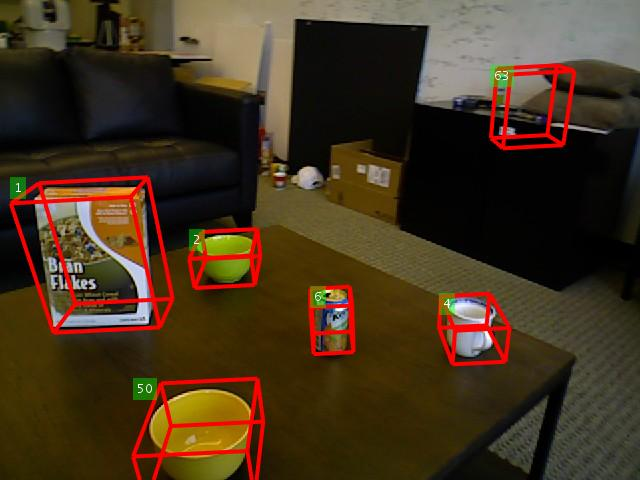}
	\caption{Frame: $500$}
	\end{subfigure}
  \begin{subfigure}[b]{0.15\textwidth}
	\includegraphics[width=\textwidth]{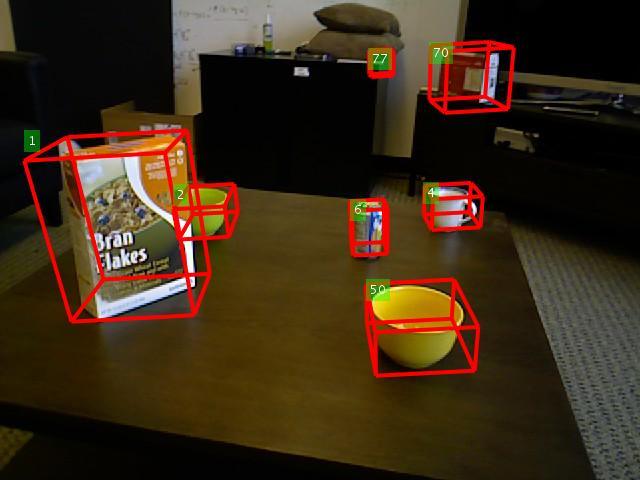}
	\caption{Frame: $600$}
	\end{subfigure}
  \begin{subfigure}[b]{0.15\textwidth}
	\includegraphics[width=\textwidth]{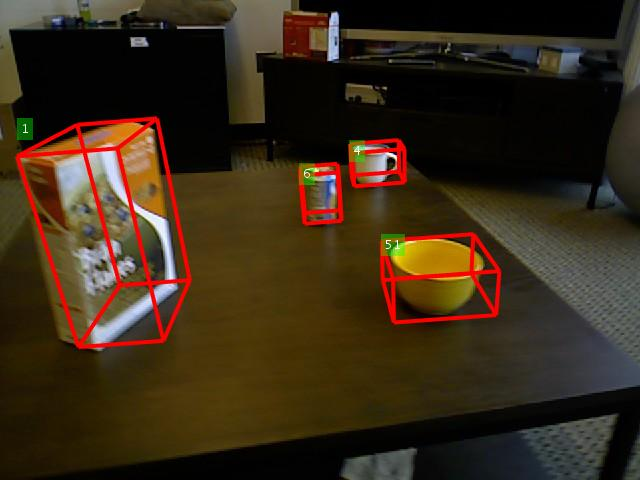}
	\caption{Frame: $644$}
	\end{subfigure}
  \begin{subfigure}[b]{0.15\textwidth}
	\includegraphics[width=\textwidth]{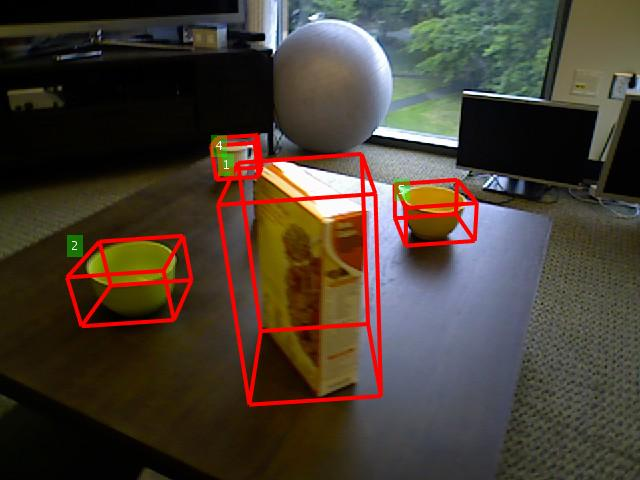}
	\caption{Frame: $742$}
	\end{subfigure}
  \caption{We show another sample scene in the UW-RGBD scene dataset for our shape matching approach. Full video can be accessed \href{https://goo.gl/WjvCsl}{here}.}
 \label{fig::uwrgbd_scene02}
\end{figure}

\end{section}

\begin{section}{Conclusion}\label{sec::conclusion}
In this paper, we have used $3$D object proposals to track and label objects in the scene. In future work, we aim to optimize our system towards real-time $3$D object tracking over even larger environments by exploring multi-scale representations for memory and computational efficiency. Ultimately, we intend to to improve the accuracy of existing SLAM techniques by integrating our system to create a semantic SLAM by treating the detected objects as fixed landmarks in the scene.
\end{section}

\bibliographystyle{IEEE_ECE}
\bibliography{thesisrefs}

\end{document}